\documentclass[conference]{IEEEtran}
\ifCLASSINFOpdf
  \usepackage[pdftex]{graphicx}
\else
\fi
\usepackage{amsmath}
\usepackage{amssymb}
\usepackage{array}

\ifCLASSOPTIONcompsoc
  \usepackage[caption=false,font=normalsize,labelfont=sf,textfont=sf]{subfig}
\else
  \usepackage[caption=false,font=footnotesize]{subfig}
\fi

\usepackage[ruled,linesnumbered]{algorithm2e}

\begin{document}
%
\title{Mapping Walls of Indoor Environment \\Using Moving RGB-D Sensor}

\author{\IEEEauthorblockN{
    Ismail\IEEEauthorrefmark{1},
    Bambang Riyanto Trilaksono\IEEEauthorrefmark{2},
    Widyawardana Adiprawita\IEEEauthorrefmark{3}}
    \IEEEauthorblockA{\IEEEauthorrefmark{1}\IEEEauthorrefmark{2}\IEEEauthorrefmark{3}School of Electrical Engineering and Informatics, 
Institut Teknologi Bandung, Bandung 40123, Indonesia\\ 
\IEEEauthorrefmark{1}School of Applied Science, 
Telkom University, Bandung 40257, Indonesia\\
Email: \IEEEauthorrefmark{1}ismail@tass.telkomuniversity.ac.id,
\IEEEauthorrefmark{2}briyanto@lskk.itb.ac.id,
\IEEEauthorrefmark{3}wadiprawita@stei.itb.ac.id}
}


%


\maketitle

\begin{abstract}
Inferring walls configuration of indoor environment could help robot ``understand``
the environment better. This allows the robot to execute a task that involves
inter-room navigation, such as picking an object in the kitchen. 
In this paper, we present a method to inferring walls configuration 
from a moving RGB-D sensor. Our goal is to combine
a simple wall configuration model and fast wall detection method
in order to get a system that
works online, is real-time, and does not need 
a Manhattan World assumption.
We tested our preliminary work, i.e. wall detection and measurement from moving RGB-D sensor,
with MIT Stata Center Dataset. 
The performance of our method is reported in terms of accuracy and
speed of execution.

\end{abstract}

%
\IEEEpeerreviewmaketitle

\section{Introduction}
A mobile robot that operates in indoor environment, 
(e.g., a service robot), needs a high-level map 
to help the robot executing a task, e.g., picking up an object in the kitchen.
Providing such a map before operation might 
be impractical because the map should be changed 
whenever the robot is placed in other environments. 

Simultaneous localization and mapping (SLAM) is a set of methods
to mapping unknown environment from
unknown robot's poses. While SLAM releases the robot
from the need of a prior map, most of SLAMs 
only provide robot with the capability of 
creating a low-level map, e.g., occupancy
grid map or feature map. 

Producing high-level map within SLAM framework
means intregating method such as object detection
and recognition into the incremental map building step.
This is shown in works such as 
\cite{castle_towards_2007}
\cite{civera_towards_2011}
\cite{rogers_simultaneous_2011}
\cite{fioraio_towards_2013}
\cite{salas-moreno_slam_2013}
\cite{galvez-lopez_real_2016}.
This approach is example of how semantic helps SLAM,
\cite{cadena_past_2016}.

Other approach is just the reverse, i.e., 
SLAM helps semantic, \cite{cadena_past_2016}.
Here, SLAM construcs a low-level map which in turn
supplied to object detection and recognition module to
building a high-level (semantic) map.

In this work, we present an approach to building
a high-level map of indoor environment from RGB-D sensor.
Specifically, we build a map of indoor structure, i.e., 
walls configuration within a floor of a building.
This kind of map helps robot to navigating itself
from room to room within the building. 

The following section describes works most related to ours.
Next to it, we show approaches we took followed
by experiment results. Finally, this paper ends by
conclusions and some works we like to pursue in the future.

\section{Related Works}
In computer vision community, indoor scene understanding usually
means extracting objects and boundaries of a room
from a single image,
\cite{dasgupta_delay_2016} 
\cite{geiger_joint_2015} 
\cite{ren_three_2016}, 
or several images,
\cite{bao_understanding_2014}.
On the contrary, in robotics, sensors are keep in motion.
While this gives robot more informations about the surrounding,
it sets a limit of time to processng those informations.
Therefore, in robotics, the ``understanding'' 
never goes into finer details. The followings are
several works on walls configuration reconstruction
with moving sensors.

Within GTSAM framework, \cite{rogers_simultaneous_2011} 
uses door signs and walls as landmarks in SLAM.
Door-signs are detected by a SVM-based classifier upon
Histogram of Oriented Gradient (HOG) features.
Walls are extracted from laser data using RANSAC.
It is important to note that the SLAM works offline,
i.e., it works after all observations data available.
This is understandable regarding the SLAM works only with
small number of landmarks (i.e., walls and door-signs)
which is insufficient to make GTSAM framework works
accurately online.

\cite{tsai_dynamic_2012} creates a model of a local indoor
environment (i.e., walls configuration) by generating
a set of hypotheses of walls perpendicular to a known ground-plane.
Each hypothesis is evaluated using observation of several points
from frame to frame basis given known camera poses.
EKF is then used to estimate posterior of each hypothesis.

\cite{furlan_free_2013} uses SLAM to generating sparse 3D point cloud
and camera pose estimate. A large number of planes 
(considered as walls, floors, or ceilings) are generated
and scored its fitness by RANSAC. Random combinations of 
available set of planes are then scored 
by particle-filter based inference engine. The highest-scored combination
are likely the correct estimate of the walls configuration.


\cite{salas_layout_2015} uses walls configuration as a help
to preventing SLAM reobserved landmarks which are occluded
by walls whenever the robot move outside a room.
This work reconstructs walls configuration by relying
on vanishing point detection on several images 
taken from several robot's poses.
These vanishing points are then projected into 3D world
to estimating planes normals. 
3D point clouds from SLAM are then aligned to these normals.
Space is then equally divided parallel to this normals. The furthest
spaces in opposite directions  (with number of points above a threshold) 
are set as walls configuration.

Our work is aimed at combining model of walls configuration
as in \cite{tsai_dynamic_2012} and the simple wall detection method
as in \cite{rogers_simultaneous_2011}. The goals are
\begin{enumerate}
  \item to drop box-shaped assumption (as in \cite{salas_layout_2015}),
  \item online (not as in \cite{rogers_simultaneous_2011}), and
  \item real-time (not just near real-time as in \cite{furlan_free_2013})
\end{enumerate}

\section{Methods}
\begin{figure}
  \centering
  \includegraphics[width=8cm]{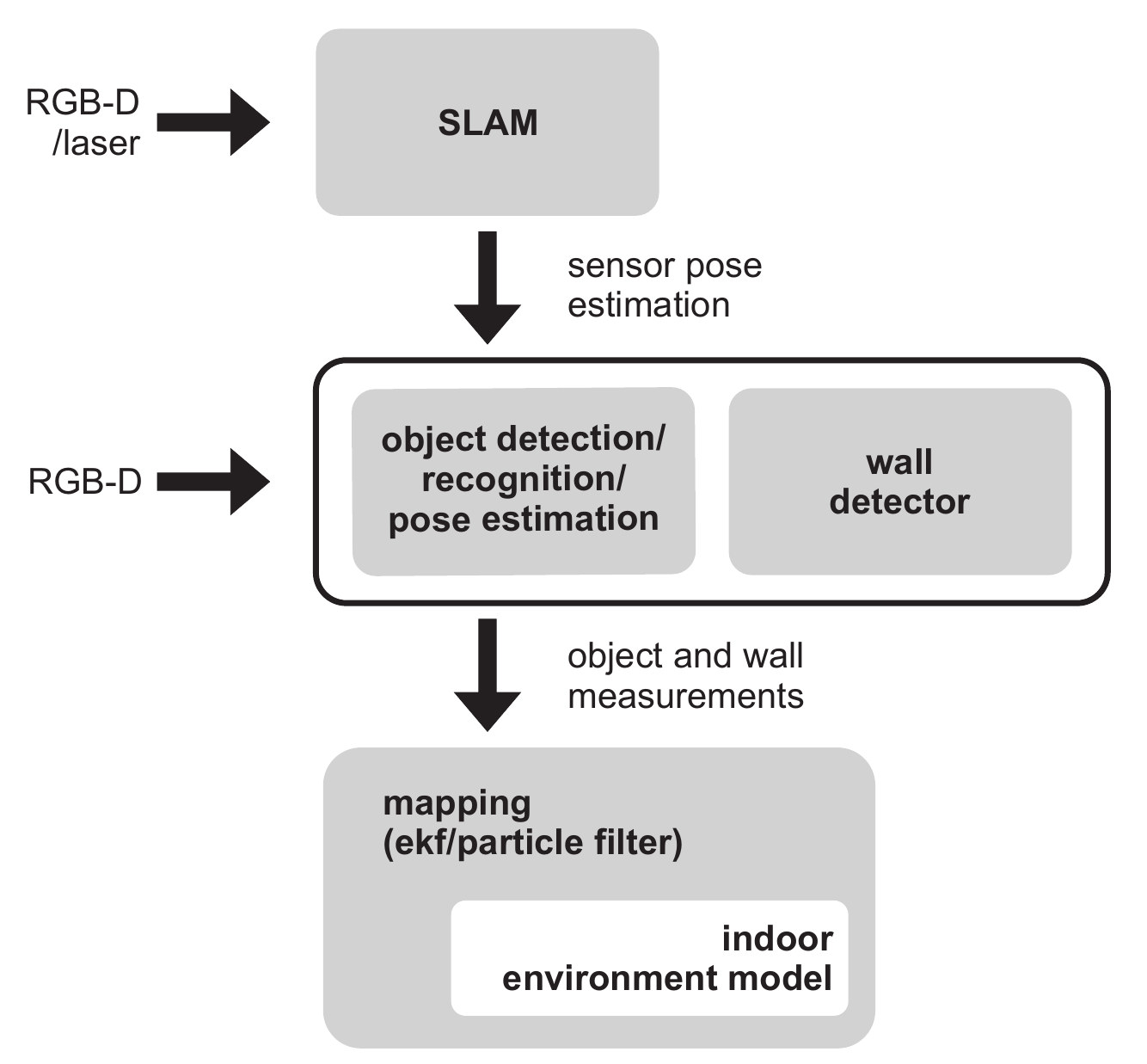}
  \caption{Block diagram of the overall system}
  \label{fig:diagram}
\end{figure}

Our target is to building a system that can map walls
and objects in indoor environment from moving RGB-D sensor.
The overall system is shown in Fig. \ref{fig:diagram}.
In this paper, we only present wall detector and 
mapping part of the system. 
As for SLAM, we used gmapping \cite{grisetti_improved_2007}
although our system should be agnostic of any kind of SLAM implementation
since we only need sensor pose estimation from it.

\subsection{Wall and Sensor Model}
\begin{figure}[!h]
  \centering
  \includegraphics[width=6cm]{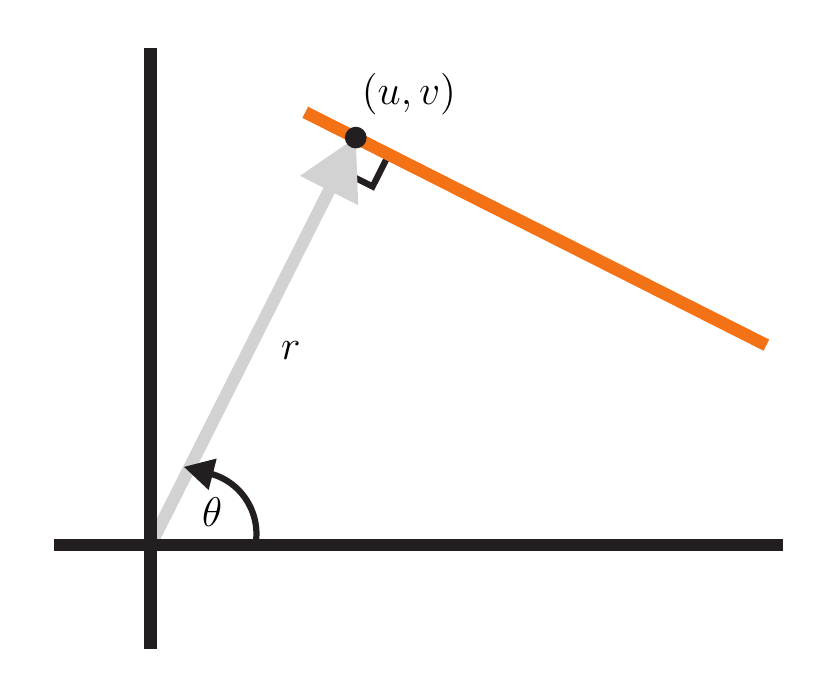}
  \caption{Model for the wall}
  \label{fig:parameter}
\end{figure}

We model a wall by a 2D line i.e., line on the ground-plane.
Further, we assume walls are perpendicular to ground-plane.
Wall is parameterized by its closest point to origin, $\mathbf{w} = (u,v)$.
(Fig. \ref{fig:parameter}).

We use the same line parameterization both in observation space
and in state space. Therefore, sensor model works like point translation/rotation
in 2D space $(c = \cos, s = \sin)$:
\begin{equation}
  \mathbf{h} = 
  \begin{bmatrix}
    u' \\ v' \\ 1 
  \end{bmatrix} =
  \begin{bmatrix}
    c(\theta) & s(\theta) & -xc(\theta) - ys(\theta) \\
    -s(\theta) & c(\theta) & xs(\theta) - yc(\theta) \\
    0 & 0 & 1
  \end{bmatrix}
  \begin{bmatrix}
    u \\ v \\ 1
  \end{bmatrix}
\end{equation}
where $(x, y, \theta)$ is robot's pose from where the wall is observed.

The inverse sensor model is
\begin{equation}
  \mathbf{h}^{-1} = 
  \begin{bmatrix}
    u \\ v \\ 1 
  \end{bmatrix} =
  \begin{bmatrix}
    c(\theta) & -s(\theta) & x \\
    s(\theta) & c(\theta) & y \\
    0 & 0 & 1
  \end{bmatrix}
  \begin{bmatrix}
    u' \\ v' \\ 1
  \end{bmatrix}
\end{equation}

Jacobian of sensor model w.r.t walls state space is
\begin{equation}
  \mathbf{H} = 
  \begin{bmatrix}
    c(\theta) & s(\theta) \\
    -s(\theta) & c(\theta)
  \end{bmatrix}
\end{equation}

\subsection{Wall Detector}
To detect walls ($\mathcal{L}$) from RGB-D Sensor, we use line fitting 
(RANSAC) on a specific row in point cloud data ($\mathcal{P}$).
We choose it to be a row in upper part of the point cloud
to avoid occlusion by objects. \texttt{wall\_detector()} 
is shown in Algorithm \ref{algo: wall-detector}.
\begin{algorithm}
  \footnotesize
  \SetKwFunction{clusterByNaN}{clusterByNaN}
  \SetKwFunction{seqLineFittingByRansac}{seqLineFittingByRansac}
  \KwIn{$\mathcal{P}$}
  \KwResult{$\mathcal{L}$}

  \Begin 
  {
      $\mathcal{C} = \varnothing$ \tiny {\tcc*[r]{line cluster}} \footnotesize
      $\mathcal{P}' \leftarrow \mathcal{P}$ \tiny {\tcc*[r]{only take a row of pointclouds}} \footnotesize
      $\mathcal{C} \leftarrow$ \clusterByNaN ($\mathcal{P}'$)\;
      $\mathcal{L} \leftarrow$ \seqLineFittingByRansac ($\mathcal{C}$)\;
      \Return{$\mathcal{L}$}
  }
  \caption {\texttt{wall\_detector()}}
  \label{algo: wall-detector}
\end{algorithm}

\begin{algorithm}
  \footnotesize
  \SetKwFunction{isFinite}{isFinite}
  \KwIn{$\mathcal{P}'$}
  \KwResult{$\mathcal{C}$}
  \Begin 
  {
      $\mathcal{C}_i = \varnothing$\; 
      \ForEach{$p$ in $\mathcal{P}'$}
      {
        \If{\isFinite ($p$)}{
          $\mathcal{C}_i \leftarrow \mathcal{C}_i + \{p\}$\;
          continue;
        } 

        $\mathcal{C} \leftarrow \mathcal{C} + \{\mathcal{C}_i\}$\;
        $\mathcal{C}_i = \varnothing$\;
      }
      \Return{$\mathcal{C}$}\;
  }
  \caption {\texttt{clusterByNan()}}
  \label{algo: clusternan}
\end{algorithm}

\begin{algorithm}
  \footnotesize
  \SetKwFunction{lineFitting}{lineFitting}
  \SetKwFunction{removeInliers}{removeInliers}
  \KwIn{$\mathcal{C}$}
  \KwResult{$\mathcal{L}$}

  \Begin 
  {
    \ForEach{$\mathcal{C}_i$ in $\mathcal{C}$}
      {
        \While{$\mathcal{C}_i \neq \varnothing$}{
          $\mathcal{L}_i$ $\leftarrow$ \lineFitting ($\mathcal{C}_i$)\;
          \removeInliers($\mathcal{C}_i$)\;
          $\mathcal{L} \leftarrow \mathcal{L} + \{\mathcal{L}_i\}$\;
        }
      }
      \Return{$\mathcal{L}$}
  }
  \caption {\texttt{seqLineFittingByRansac()}}
  \label{algo: seqlinefittingbyransac}
\end{algorithm}

RANSAC returns wall in parameter $m$ and $c$. 
It is straightforward to change it to 
point closest to origin.
\begin{equation}
  \begin{aligned}
    u &= \frac{-m c}{m^2+1} \\
    v &= \frac{c}{m^2+1}
  \end{aligned}
\end{equation}

\subsection{Mapping Wall with EKF}
Our goal is to calculating probability density function of
walls given observations ($\mathbf{z}_{1:t}$) and robot's poses ($\mathbf{x}_{1:t}$):
\begin{equation}
  p(\mathbf{w}_{1:M} | \mathbf{x}_{1:t}, \mathbf{z}_{1:t})
  \label{eq:posterior}
\end{equation}

\begin{figure}
  \centering
  \includegraphics[width=8cm]{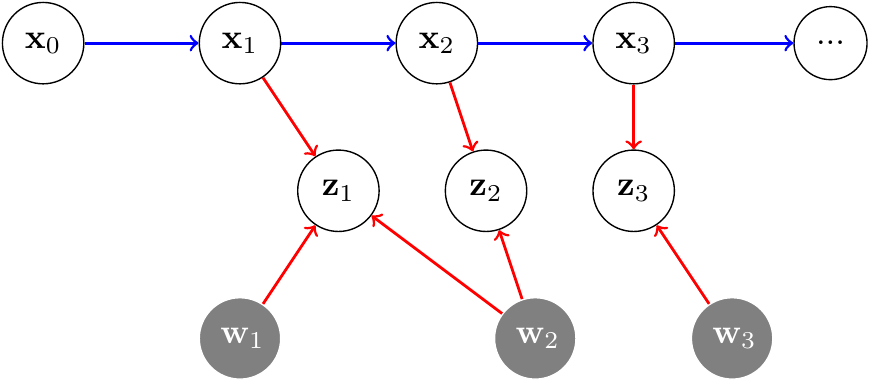}
  \caption{Dynamic Bayesian Network for walls mapping problem. Shaded nodes indicate unknown variables.}
  \label{fig:dbn}
\end{figure}

From Dynamic Bayesian Network (Fig. \ref{fig:dbn})
and using concept
of conditional independence, Equation \ref{eq:posterior}
could be factored as
\begin{equation}
  p(\mathbf{w}_{1:M}|\mathbf{x}_{1:t}, \mathbf{z}_{1:t}) = \prod_{i=1}^M p(\mathbf{w}_i|\mathbf{x}_{1:t}, \mathbf{z}_{1:t})
\end{equation}

It means we could calculate the posterior of
landmarks from product of posteriors for individual landmark.

Within Recursive Bayesian Framework, with the Gaussian
assumption of posterior ($\mathbf{w} \sim \mathcal{N}(\mathbf{\bar{w}},\Sigma_\mathbf{w})$), 
each individual landmark
could be estimated using Extended Kalman Filter.
\begin{equation}
  \begin{aligned}
    \mathbf{w}_i &\leftarrow \mathbf{w}_i + \mathbf{K} (\mathbf{z}_k - \mathbf{h}) \\
    \Sigma_\mathbf{w} &\leftarrow (\mathbf{I} - \mathbf{K}\mathbf{H})\Sigma_\mathbf{w} \\
  \end{aligned}
\end{equation}
with $\mathbf{K}$ is Kalman Gain
\begin{equation}
  \mathbf{K} = \Sigma_\mathbf{w}\mathbf{H}^T(\mathbf{R} + \mathbf{H}\Sigma_\mathbf{w}\mathbf{H}^T)^{-1}
\end{equation}
$\mathbf{R}$ is covariance of measurement noise. 

\subsection{Data Association}
We use maximum likelihood to find whether a given
observation is coming from the already observed wall.
The likelihood function is defined
\begin{equation}
  \begin{aligned}
    L &= p(\mathbf{z}_{i,k}|\mathbf{w}_i,\mathbf{x}_k) \\
    &\propto \exp(-0.5 \times (\mathbf{z}-\mathbf{h})^T\Sigma_\mathbf{w}^{-1}(\mathbf{z}-\mathbf{h}))
  \end{aligned}
\end{equation}

Because there are a small number of walls in environment,
we use exhaustive search to do data association.
A sensor observes a new wall whenever the maximum likelihood
falls below a certain threshold.

\section{Experiments}
We use ROS package of grid mapping as an underlying SLAM system.
We tested our system in MIT Stata Center Dataset \cite{fallon_mit_2013}.

\begin{figure*}
  \centering
  \includegraphics[width=6cm]{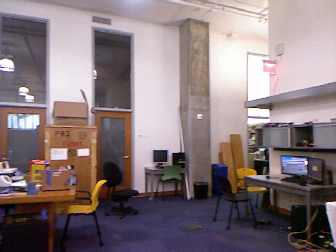}
  \includegraphics[width=6cm]{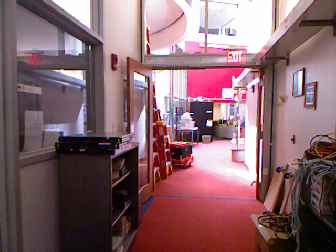} \\
  \includegraphics[width=6cm]{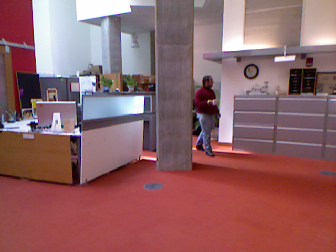}
  \includegraphics[width=6cm]{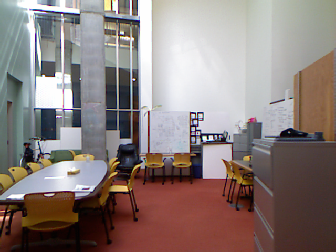}
  \caption{Example frames from MIT Stata Center Dataset}
  \label{fig:mit}
\end{figure*}

\begin{figure*}[!h]
  \centering
  \includegraphics[width=18cm]{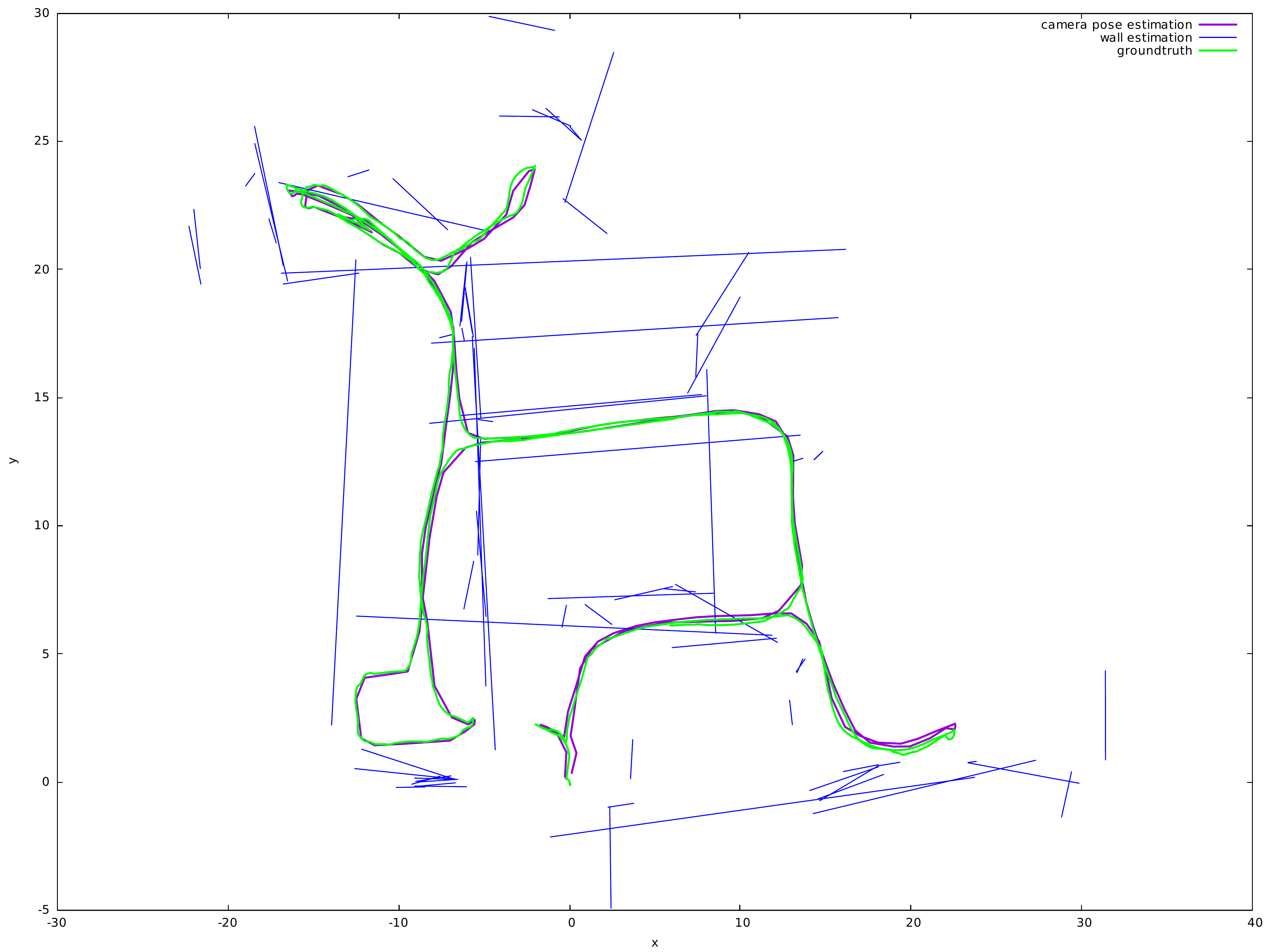}
  \caption{Result with robot's trajectory estimation using gmapping.}
  \label{fig:result}
\end{figure*}

\begin{figure}
  \centering
  \includegraphics[width=8cm]{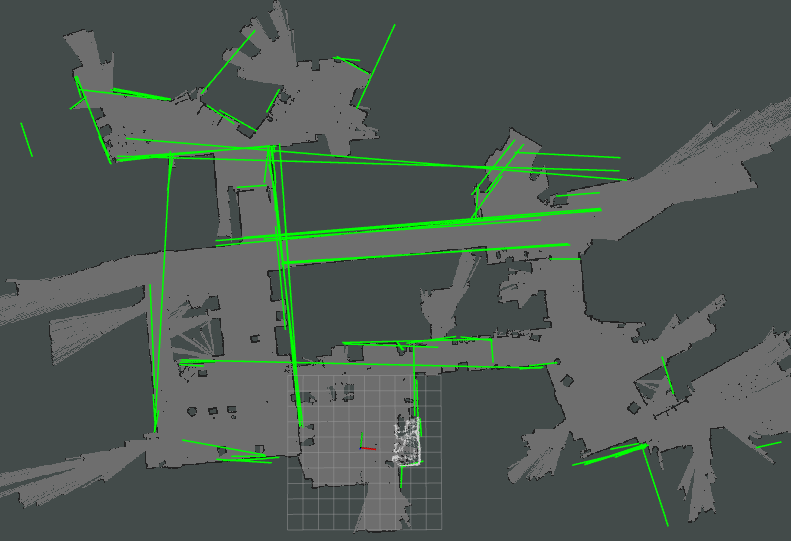}
  \caption{Result overlaid on occupancy grid from gmapping}
  \label{fig:resultwithgrid}
\end{figure}

MIT Stata Center Dataset is a very challenging dataset because 
it consists of indoor environment with irregular
shape of rooms. There are numerous rooms without clear boundaries 
and walls with large windows (Fig. \ref{fig:mit}). This frequently makes 
wall detector fail to detect walls. Furthermore, the room sometimes
have a large dimension which makes RGB-D sensor fail to reach
the furthest side of its wall. This gives rather scarce wall detection rate
as seen in Fig. \ref{fig:result} and Fig. \ref{fig:resultwithgrid}.

\begin{figure*}
  \centering
  \includegraphics[height=5.5cm]{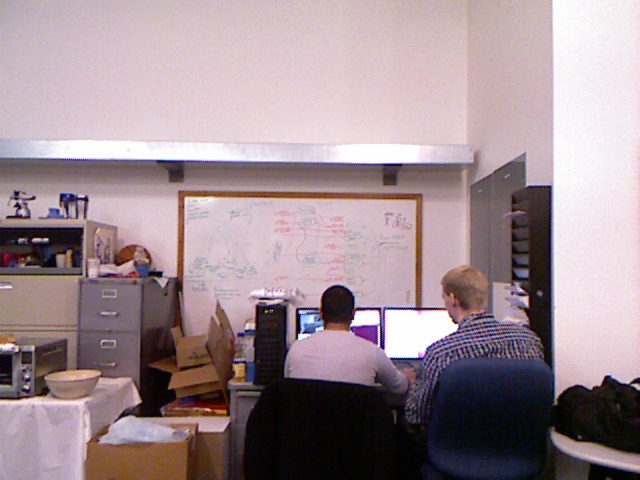}
  \includegraphics[height=5.5cm]{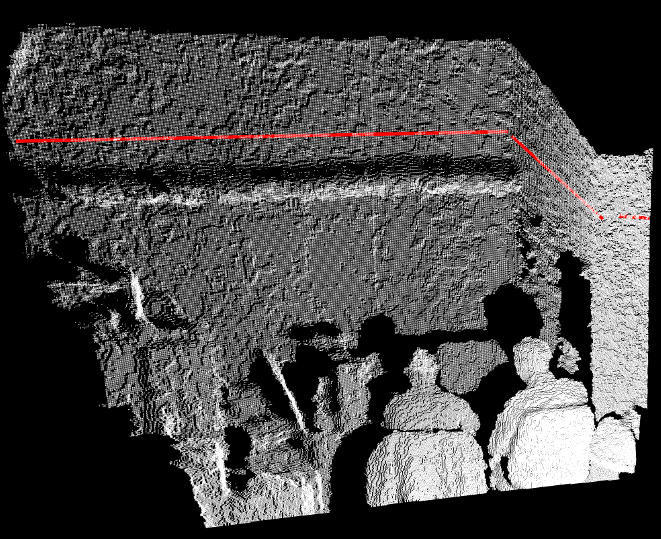}
  \caption{Wall detector algorithm successfully detects a wall in relatively simple scene}
  \label{fig:success}
\end{figure*}

\begin{figure*}
  \centering
  \includegraphics[height=6cm]{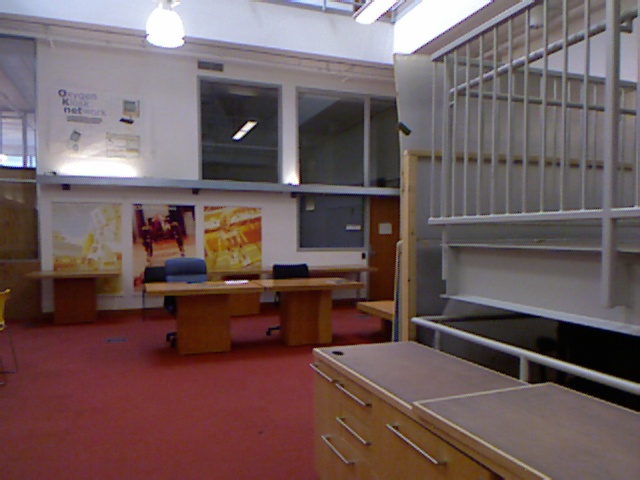}
  \includegraphics[height=6cm]{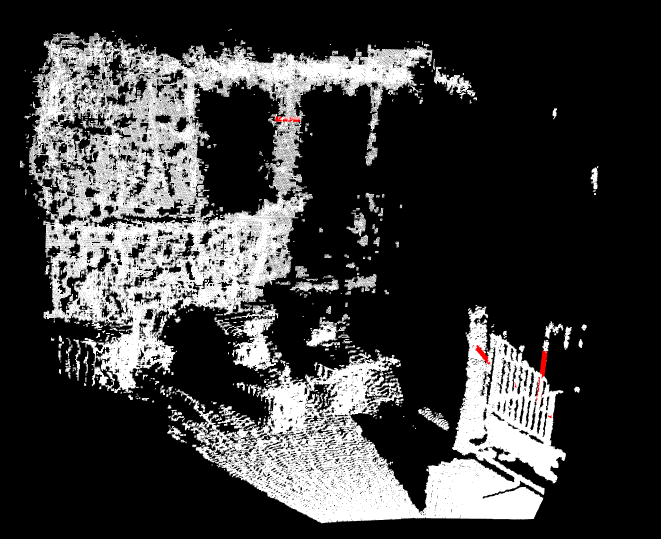}
  \caption{Challenging frame: the wall is relatively far from sensor making greater noise in pointclouds. Wall detector failed to detect large portion of the wall in the left side of the frame.}
  \label{fig:failed}
\end{figure*}

\begin{figure}
  \centering
  \includegraphics[width=8cm]{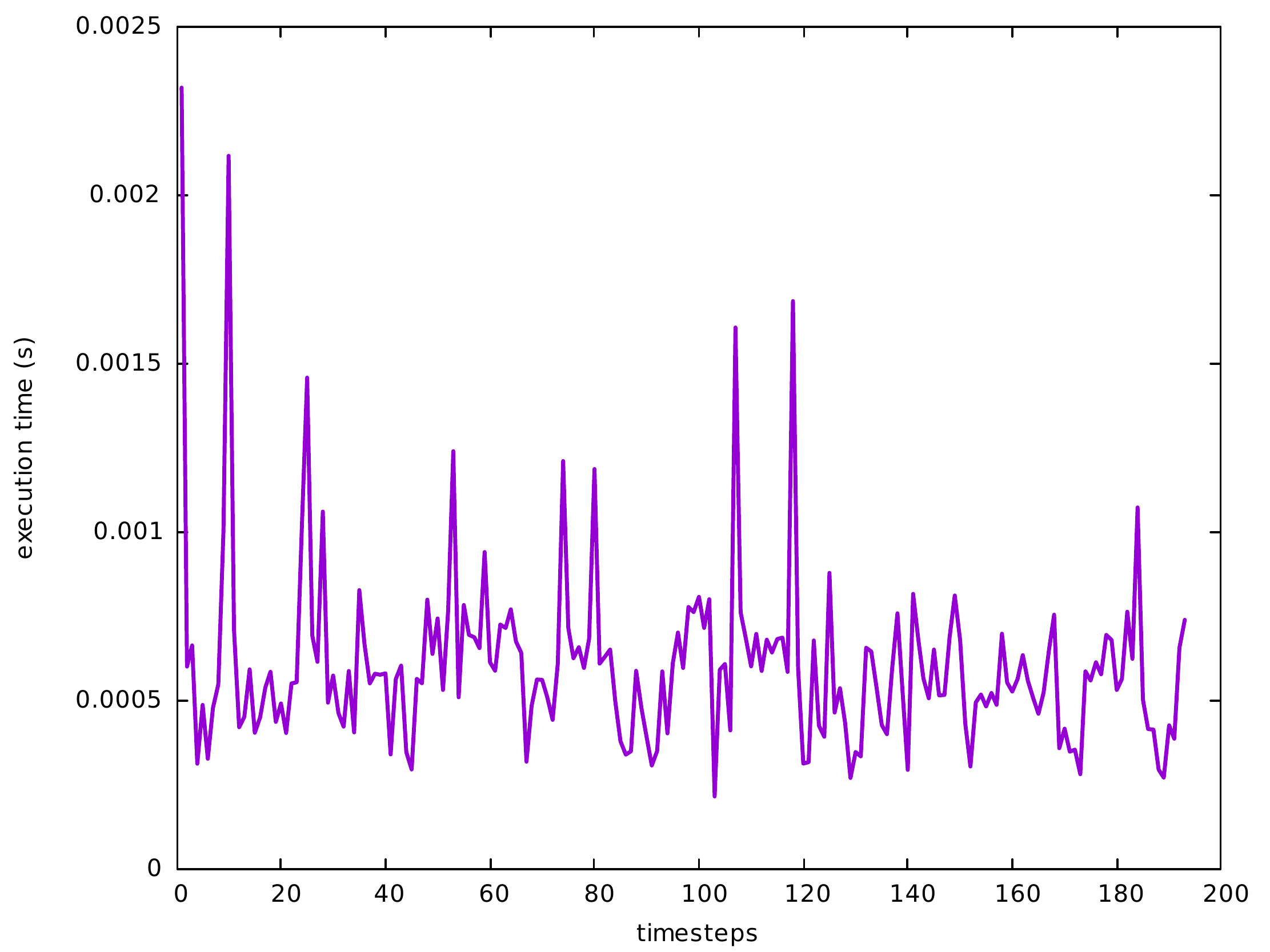}
  \caption{Execution time of \texttt{wall\_detector()} algorithm.}
  \label{fig:twalldetection}
\end{figure}

Our wall detection method is simple but it has acceptable
detection performance. Fig. \ref{fig:twalldetection} shows
detection performance of our method. Fig. \ref{fig:success}
and Fig. \ref{fig:failed}
shows result of wall detector algorithm to two frames
in dataset.

\begin{figure}
  \centering
  \includegraphics[width=8cm]{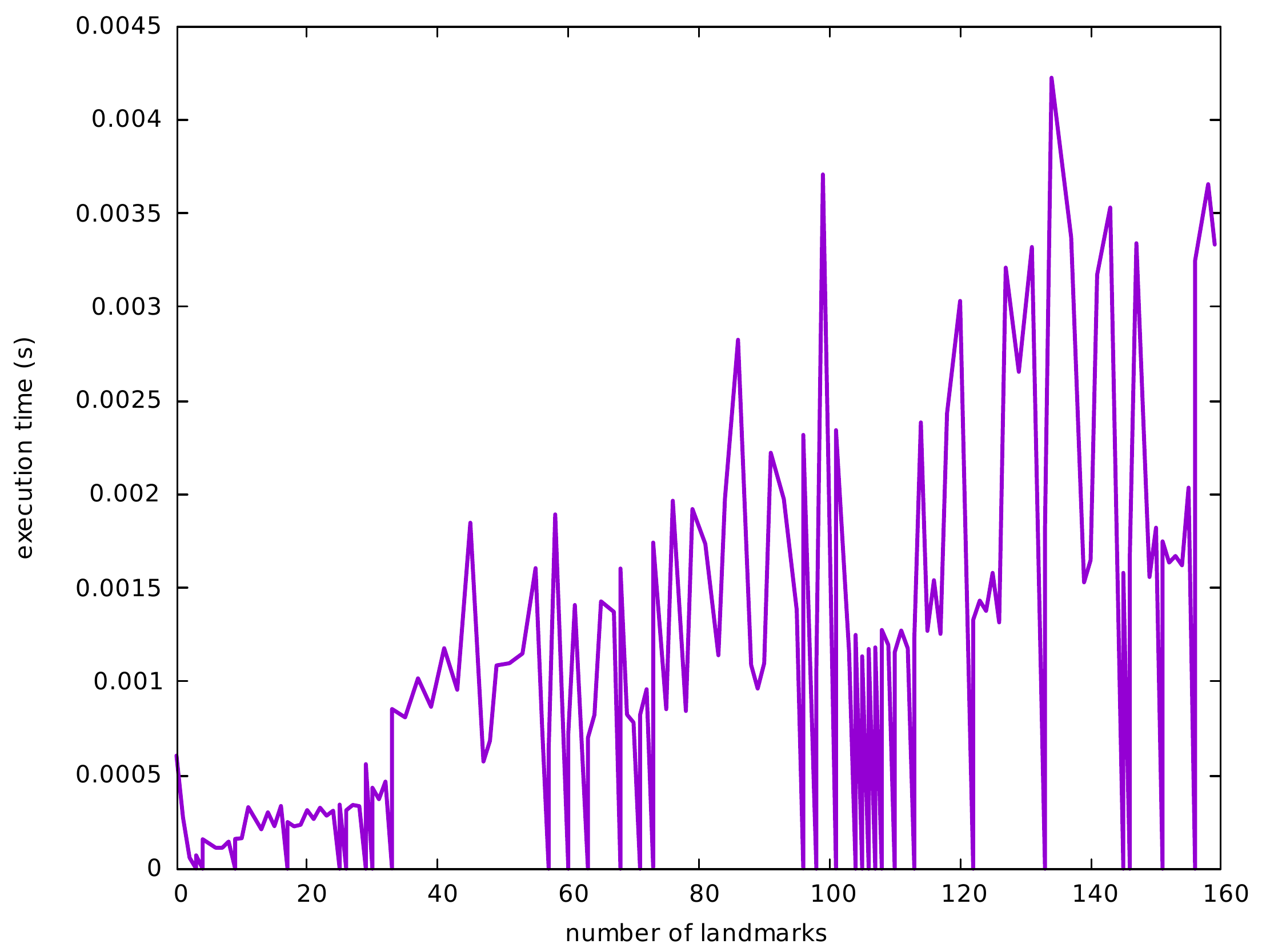}
  \caption{Data association vs number of landmarks}
  \label{fig:tdataassociation}
\end{figure}

Other critical component in our system is data association.
Although we used exhaustive search, the number of walls never
grows very large. Fig. \ref{fig:tdataassociation} shows performance
of the method.

\begin{figure}
  \centering
  \includegraphics[width=8cm]{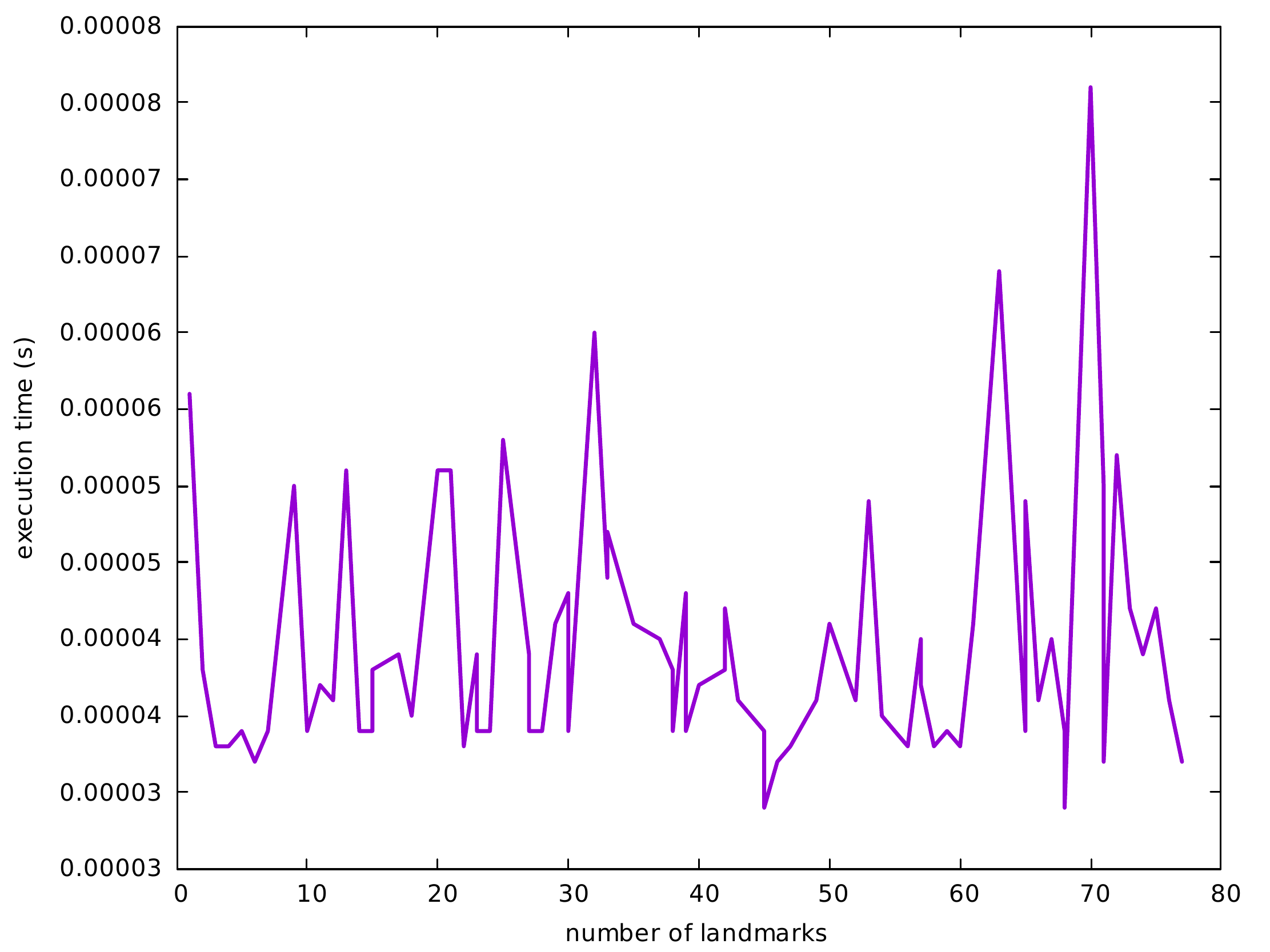}
  \caption{Execution time of EKF update vs number of landmarks}
  \label{fig:tekf}
\end{figure}

In EKF-based SLAM, updating the covariance matrix would
be the most time consuming process. Here, our EKF is in 
small and constant dimension.
It does not depend on the number
of landmarks (Fig. \ref{fig:tekf}).

\begin{figure}
  \centering
  \includegraphics[width=8cm]{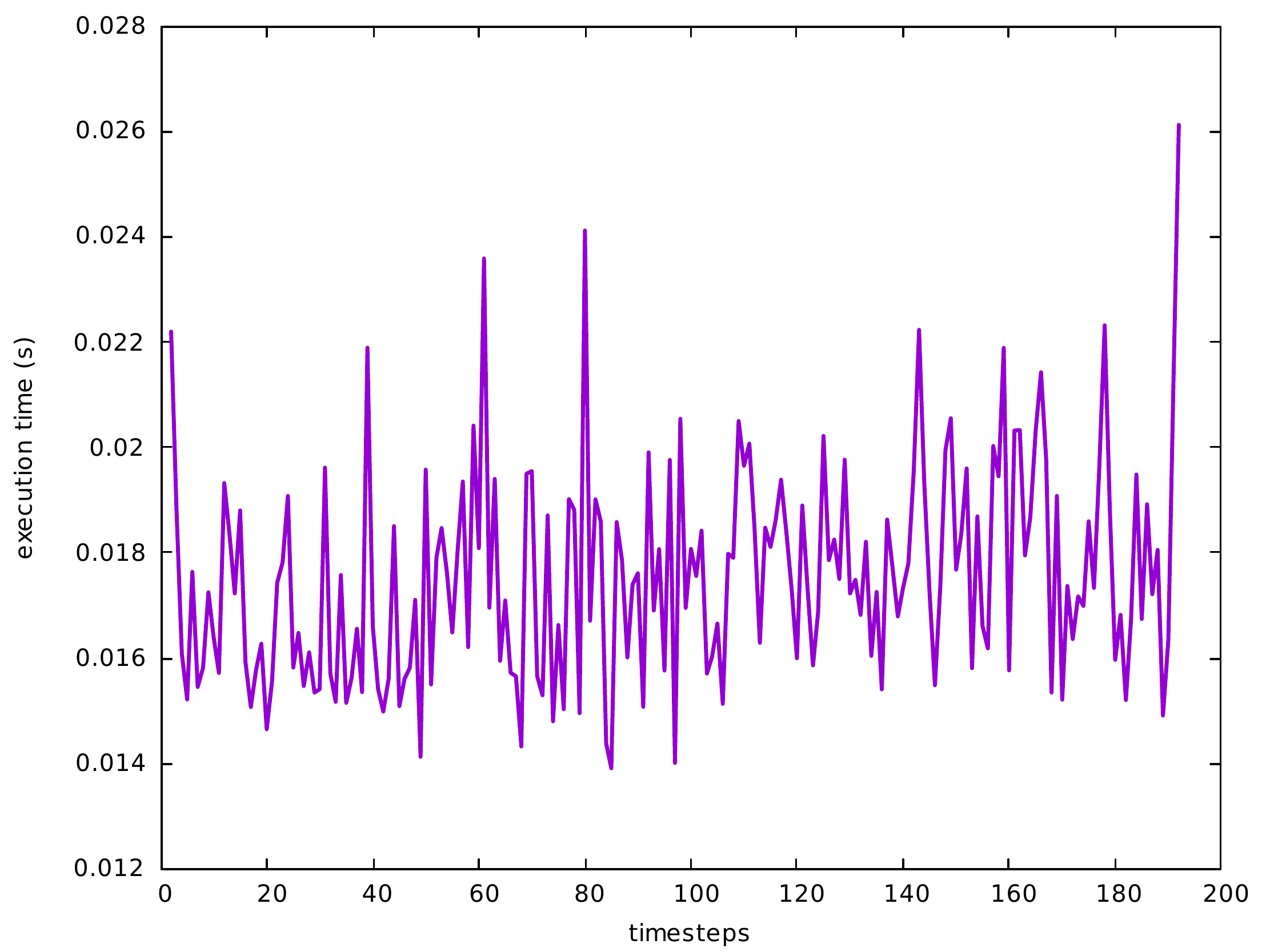}
  \caption{Execution time of our method (without gmapping time)}
  \label{fig:all}
\end{figure}

Overall, Fig. \ref{fig:all} show the execution time of our method
plotted in timesteps. This excludes time consumed by gmapping
as underlying component of our method. For the mapping result,
it is shown in Fig. \ref{fig:result}.

In our experiments, we use software and hardware
as listed in Table \ref{tab:wares}.

\begin{table}[!h]
    \centering
    \caption{Softwares and hardwares used in experiments}
    \label{tab:wares}
    \begin{tabular}{p{3cm} | p{4cm}}
        \textbf{Soft/Hardwares} & \textbf{Specifications} \\  \hline
        Dataset & MIT Stata Center Dataset\\
        Processor& Intel i5-3330 3GHz processor \\
        RAM & 16GB \\
        Pointclouds library & pcl \\
        Software framework & ROS \\
    \end{tabular}
\end{table}

\section{Conclusion and Future Work}
We seek to building a method to mapping walls in indoor environment which 
\begin{enumerate}
  \item has a simple model to represents walls configuration
  \item has the capability of working online
  \item has no assumption about the shape of the room
\end{enumerate}

In this paper, we showed our preliminary work on wall detector from moving RGB-D sensor 
and showed the performance in terms of speed of execution.

Our future works would be to improve the accuracy of 
wall detection and implementing model of walls configurations.
Further, we seek to develop a method to also map objects in environment.



\IEEEtriggeratref{11}


%




\bibliographystyle{IEEEtran}
\bibliography{IEEEabrv,/home/ism/data/paper/mylib.bib}

\end{document}